\documentclass[conference]{templates/ieeeconf}
\IEEEoverridecommandlockouts

\let\labelindent\relax
\usepackage{enumitem}
\usepackage[noadjust]{cite}
\usepackage{booktabs}
\usepackage[utf8]{inputenc} 
\usepackage[T1]{fontenc}    
\usepackage{graphicx}
\usepackage{tabularx,longtable,multirow,caption}
\usepackage{graphics} 
\usepackage[style=base]{subcaption}
\usepackage{url}            
\usepackage{amsfonts}       
\usepackage{mathrsfs}
\usepackage{amsthm}
\usepackage{wrapfig}

\usepackage{color}
\usepackage{pgf, tikz, pgfplots}
\usetikzlibrary{shapes, arrows, automata}
\usepackage[font = small]{caption} 
\usepackage{amsfonts, amssymb}
\usepackage{amsmath}
\usepackage{hyperref}      

\input{templates/pennColors.sty}

\usepackage{algorithm}
\usepackage[noend]{algorithmic}
\newcommand\NoDo{\renewcommand\algorithmicdo{}}

\newcommand\NoThen{\renewcommand\algorithmicthen{}}

\begin{document}

\title{\LARGE \bf
Accelerating Multi-Agent Planning Using Graph Transformers\\ with Bounded Suboptimality
}

\author{Chenning Yu$^{*, 1}$, Qingbiao Li$^{*, 2}$, Sicun Gao$^{1}$, Amanda Prorok$^{2}$
\thanks{*Equal contribution. }%
\thanks{$^{1}$Computer Science and Engineering Department, University of California San Diego. {\tt\small <chy010, sicung>@ucsd.edu}}%
\thanks{$^{2}$Department of Computer Science and Technology, University of Cambridge. {\tt\small <ql295, asp45>@cam.ac.uk}}} %

\maketitle

\begin{abstract}

Conflict-Based Search is one of the most popular methods for multi-agent path finding. Though it is complete and optimal, it does not scale well. Recent works have been proposed to accelerate it by introducing various heuristics. However, whether these heuristics can apply to non-grid-based problem settings while maintaining their effectiveness remains an open question. In this work, we find that the answer is prone to be no. To this end, we propose a learning-based component, i.e., the Graph Transformer, as a heuristic function to accelerate the planning. The proposed method is provably complete and bounded-suboptimal with any desired factor. We conduct extensive experiments on two environments with dense graphs. Results show that the proposed Graph Transformer can be trained in problem instances with relatively few agents and generalizes well to a larger number of agents, while achieving better performance than state-of-the-art methods.

\end{abstract}

\begin{figure*}[!t]
  \centering
  \includegraphics[width=1.0\textwidth]{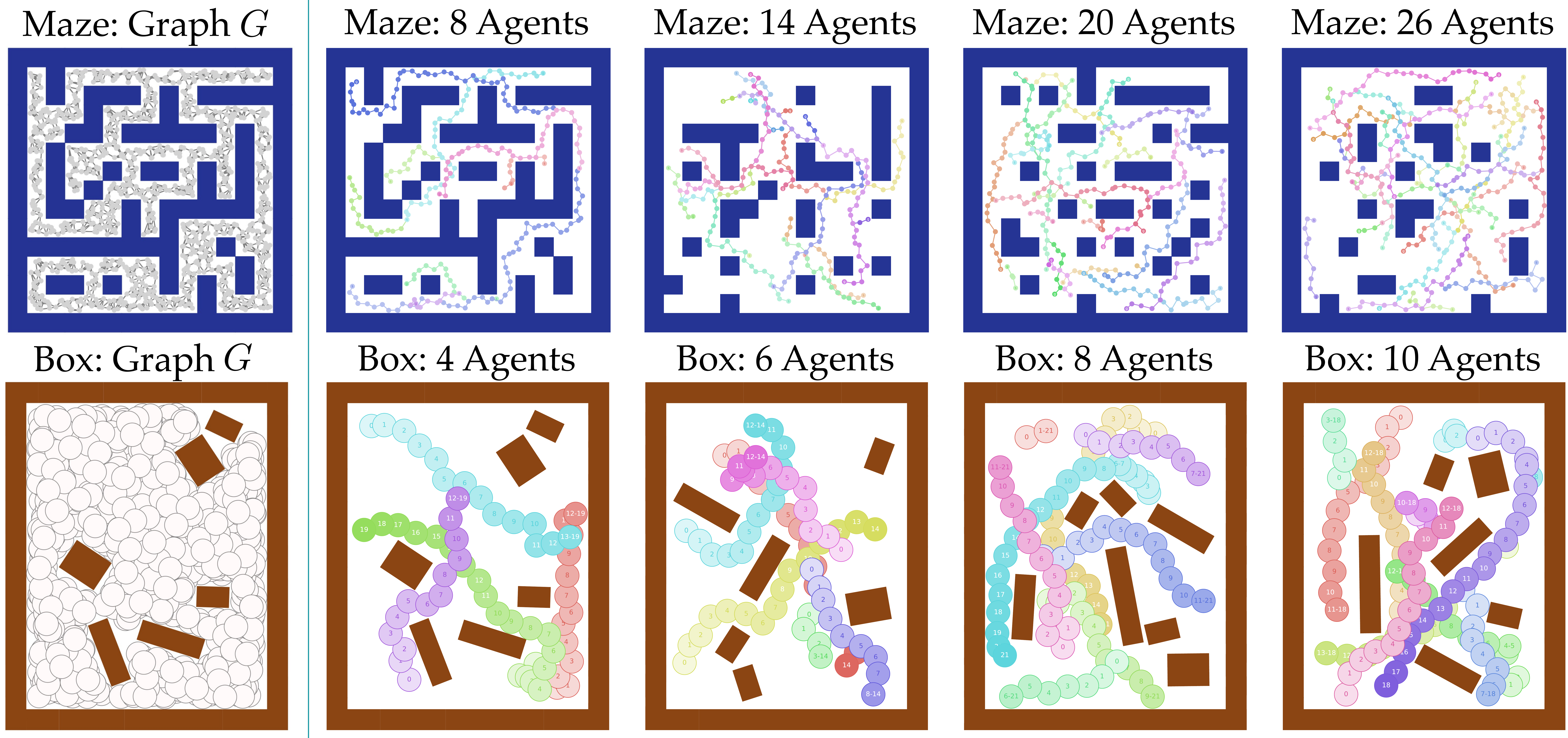}
  \caption{\textbf{Left:} Examples of our graph-based MAPF instances. To construct the graph, we sample vertices randomly from the free space and connect them with collision-free edges. \textbf{Right:} Problem instances that our approach solves while other baselines fail. Different colors represent the trajectories of different agents. Vertices in the same trajectory have deeper colors if their respective time steps are later.}
  \label{fig: problem_instance}
  \vspace{-10pt}
\end{figure*}


\section{Introduction}

Multi-Agent Path Finding (MAPF) is central to many multi-agent problems. The solution to MAPF is to generate collision-free paths guiding agents from their start positions to designated goal positions.  MAPF has practical applications in item retrieval in warehouses~\cite{enright:2011}, mobility-on-demand services~\cite{prorok2017privacy}, surveillance~\cite{grocholsky2006cooperative} and search and rescue~\cite{kumar2017opportunities}. 

Conflict-Based Search (CBS) is one of the most popular planners for MAPF~\cite{sharon_Conflictbased_2015}. It is provably complete and optimal. However, solving MAPF optimally is NP-hard~\cite{nphard1,nphard2}. Consequently, CBS suffers from scalability, as the search space grows exponentially with the number of agents.
Bounded-suboptimal algorithms~\cite{barer2014suboptimal,li2021eecbs} guarantee a solution that is no larger than a given constant factor over the optimal solution cost. Though these methods often run faster than CBS for grid-based MAPF instances, their effectiveness remains an open question for non-grid-based problem settings, wherein agents can move in an arbitrary continuous domain. In addition, since most of these heuristics rely heavily on collision checking for conflicts, their computational costs may become considerable when the graphs are dense.

Recently, learning-based methods have shown their potential in solving MAPF tasks efficiently~\cite{Qingbiao2019,huang2021learning,huang2021learningb,sartoretti_PRIMAL_2019}, which offload the online computational burden into an offline learning procedure. Yet, we find nearly none of them address graph-based MAPF settings. Therefore, the main interest of this work is to explore whether and how a learning-based method could accelerate MAPF planners in non-grid-based settings, especially for dense graphs.

\textbf{Contributions.}
We propose to use a Graph Transformer as a heuristic function to accelerate Conflict-Based Search (CBS) in a non-grid setting. Similar to previous works~\cite{huang2021learningb}, by introducing focal search to CBS, our framework guarantees both the completeness and bounded-suboptimality of the solution. Our contributions are as follows:

\begin{itemize}[leftmargin=*]
\item We propose a novel architecture, i.e., the Graph Transformer, which leverages the underlying structure of the MAPF problem. The proposed architecture has several desired properties, e.g., dealing with an arbitrary number of agents, making it a natural fit for the MAPF problem. To our knowledge, our work is one of the first works to introduce a learning component to MAPF problems under {\em non-grid-based} problem settings.
\item We design a novel training objective, i.e., Contrastive Loss, to learn a heuristic that ranks the search nodes. Unlike~\cite{huang2021learningb}, our loss can be directly optimized without introducing an upper bound, which is suitable for deep learning.
\item We demonstrate the generalizability of our model by training with relatively few agents and testing in unseen instances with larger agent numbers. Results show that our approach can accelerate CBS significantly while ECBS, using handcrafted heuristics~\cite{barer2014suboptimal}, fails.
\end{itemize}



{\bf \em Related Work.} Leading MAPF planners mainly include three types: optimal planners, bounded-suboptimal planners, and unbounded-suboptimal planners. Optimal planners include BCP solvers~\cite{MILP1, MILP2}, and Conflict-Based Search (CBS)~\cite{sharon_Conflictbased_2015}, followed by its variants, e.g., CBSH~\cite{CBSH} and CBSH2~\cite{CBSH2}. Bounded-suboptimal planners are another line of work with better scalability while guaranteeing completeness and bounded-suboptimality. Representative works include EPEA*~\cite{epea},  A* with operator decomposition~\cite{AOD}, M*~\cite{mstar}, ECBS~\cite{barer2014suboptimal}, and EECBS~\cite{li2021eecbs}. Last but not least, there are unbounded-suboptimal planners that aim to aggressively accelerate the planning. Examples include Prioritized Planning \cite{van2005prioritized}, ORCA~\cite{vandenberg_Reciprocal_2008}, Push-and-Swap~\cite{luna2011push}, and Parallel Push-and-Swap~\cite{sajid2012multi}. These works can find solutions fast, but do not guarantee the solution quality~\cite{vcap2015prioritized, van2005roadmap}.

Recently, learning-based methods were introduced to solve multi-agent tasks efficiently, using imitation learning~\cite{Qingbiao2019,li2020message} and reinforcement learning~\cite{sartoretti_PRIMAL_2019,wang2020mobile,https://doi.org/10.48550/arxiv.2210.09378}. These end-to-end methods are often good at memorizing the patterns that are seen during training, which could save significant online computation when deployed to similar tasks. However, it is hard to ensure their completeness. To this end, several works have been proposed to train a learning-based heuristic and use it to guide the tree search~\cite{khalil2017learning,DBLP:journals/corr/abs-2210-08408,song2018learning,yu2021reducing}. 
To accelerate CBS and ECBS, Huang et al. \cite{huang2021learning,huang2021learningb} use Supported Vector Machines (SVM) to bias the search, and train it via imitation learning. Our work is not only one of the first works that apply ML to {\em non-grid-based} problem settings, but also one of the first works that apply {\em deep learning} to MAPF with completeness and bounded-optimality guarantees.

\section{Problem Formulation}\label{sec:problem_formulation}

We study Multi-Agent Path Finding (MAPF) in the 2D continuous space $\mathcal{C}\subseteq \mathbb{R}^2$. The configuration space $\mathcal{C}$ consists of a set of obstacles $\mathcal{C}_{obs} \subseteq \mathcal{C}$ and free space $\mathcal{C}_{free}: \mathcal{C} \setminus \mathcal{C}_{obs}$. Note that $\mathcal{C}_{obs}$ could be different from what appears in the workspace, since it also considers the geometric shape of the agent, which may not solely be a point mass. 

A random geometric graph $G=\langle V, E\rangle$ is sampled from the space. Every sampled vertex $v\in V$ is collision-free, i.e., $v\in \mathcal{C}_{free}$. A directed edge $e\in E: (v_i \rightarrow v_j)$ connects $v_i$ to $v_j$, if (i) $v_j$ is one of the neighbors of $v_i$, and (ii) the edge is collision-free, i.e., $e\subseteq \mathcal{C}_{free}$. The neighbor set can be defined as the $r$-radius or $k$-nearest neighbors.

Suppose there are $M$ agents on this graph $G$. Each agent $i$ occupies a region $\mathcal{R}(q)\subseteq \mathcal{C}$, associated with a vertex $q\in V$. We assign a start vertex $s_i$ and a goal vertex $g_i$ to each agent $i$. We denote the path of agent $i$ as $\sigma_i: \{v_i^t\}_{t\in[1\cdots T_i]}$, where $T_i\in \mathbb{Z}_{>0}$, and agent $i$ is on vertex $v_i^t$ at time step $t$. We denote $e_i^t$ as the edge $(v_i^t \rightarrow v_i^{t+1})$ that traverses from $v_i^t$ to $v_i^{t+1}$ in 1 timestep.

\noindent{\bf Problem Description.} We consider a tuple $(G, \mathcal{S}, \mathcal{G}, \mathcal{C}, \mathcal{R})$ as a problem instance of MAPF, where $\mathcal{S}: \{s_i\}_{i\in[1\cdots M]}$ and $\mathcal{G}: \{g_i\}_{i\in[1\cdots M]}$ are the start and goal vertices. A conflict-free solution $\{\sigma_i\}_{i\in[1\cdots M]}$, should satisfy the following objectives, given arbitrary time $t$ and pair of agents $i, j$~\cite{CTRM}:

\noindent{\bf (Endpoint)} $v_i^0=s_i \wedge v_i^{T_i}=g_i$

\noindent{\bf (Obstacle)} $v_i^t \in \mathcal{C}_{free} \wedge e_i^t \subseteq \mathcal{C}_{free}$

\noindent{\bf (Inter-agent)} $\mathcal{R}(q_i) \cap \mathcal{R}(q_j) = \emptyset, \forall q_i \in e_i^t, q_j \in e_j^t$

\noindent{\bf Note:} If $t \geq T_i$, we assume the corresponding $v_i^t$ is equal to $v_i^{T_i}$. This means that the agent will stay at the goal starting from time step $v_i^{T_i}$.

\noindent{\bf Solution Quality.} We assume each edge requires 1 time step to traverse. The quality of the solution is measured by the sum of travel times (flowtime): $\sum_{i\in[1\cdots M]} T_i$.

\section{Background: Conflict-Based Search with Biased Heuristics}

In this section, we first introduce Conflict-Based Search (CBS), an optimal multi-agent planner~\cite{sharon_Conflictbased_2015}. Then we introduce {\em focal search}~\cite{pearl1982studies,ghallab1983efficient}, which incorporates the biased heuristic into the CBS framework, while preserving the guarantees of bounded-suboptimality and completeness~\cite{barer2014suboptimal}.

\subsection{Conflict-Based Search}

Conflict-Based Search (CBS) is an optimal bi-level tree search algorithm of MAPF. The high-level planner aims to solve inter-agent conflicts, while the low-level planner aims to generate optimal individual paths. Here we denote an inter-agent conflict as $(i,j,t,v_i^{t-1}, v_j^{t-1}, v_i^{t}, v_j^{t})$, which implies two edges, $(v_i^{t-1}\rightarrow v_i^{t})$ and $(v_j^{t-1}\rightarrow v_j^{t})$, dissatisfy the inter-agent objective mentioned in Section \ref{sec:problem_formulation}.

The high-level planner maintains a tree and decides which search node to expand in a best-first manner. To this end, each search node $N$ stores the following information:

(1) A set of constraints $\mathcal{T}(N)$. A constraint $(i, v, t)$ indicates that agent $i$ should not traverse to graph vertex $v$ at time $t$.

(2) A solution $\sigma(N)$: $\{\sigma_i\}_{i\in[1\cdots M]}$. The solution satisfies the endpoint and obstacle objective, but may or may not satisfy the inter-agent objective. In addition, the solution should obey the constraints $\mathcal{T}(N)$, i.e., $\forall i, \forall v_i^t\in \sigma_i, (i, v, t)\not \in \mathcal{T}(N)$.

(3) The cost of the solution $c(N)$. The high-level planner prioritizes which search node to expand based on this metric.

On the high level, CBS first creates a root search node with no constraints, then keeps selecting a search node and expanding it. A search node $N^*$ is selected if it is a leaf node with the lowest cost. CBS then checks whether the solution $\sigma(N^*)$ has an inter-agent conflict. If there is no conflict, $\sigma(N^*)$ will be returned as the final result. Otherwise, CBS chooses the first conflict $C$: $(i,j,t,v_i^{t-1}, v_j^{t-1}, v_i^{t}, v_j^{t})$, and splits it into two constraints $C_1$: $(i,v_i^t,t)$ and $C_2$: $(j,v_j^t,t)$. Two child search nodes $N_1$ and $N_2$ are then generated, with constraints $\forall i=1,2,\mathcal{T}(N_i)=\mathcal{T}(N^*)\cup \{C_i\}$ respectively. Then an optimal low-level planner, e.g., A*~\cite{astar}, is called by each child search node, which replans the path for each affected agent and records the respective solution and cost. CBS guarantees completeness and optimality, since both the high-level and low-level planners are performing best-first search~\cite{sharon_Conflictbased_2015}.

\subsection{Incorporating Biased Heuristics using Focal Search}

CBS is an optimal planner, but it does not scale well even for grid-based problems settings. To improve the scalability, focal search~\cite{pearl1982studies,ghallab1983efficient} was introduced by previous works, e.g., Bounded CBS (BCBS) and Enhanced CBS (ECBS)~\cite{barer2014suboptimal}. Here we describe a simplified version of BCBS.

We present the pseudocode of focal search in Algorithm \ref{alg:focal_search}. Focal search introduces the focal set to the CBS framework. A focal set ($Focal$) maintains a fraction of the leaf search nodes in the CBS tree (i.e., $Open$). We denote $LB$ as the lowest solution cost in leaf search nodes. All the leaf search nodes that satisfy a near-optimal solution quality $c\leq w\cdot LB$ will be added to $Focal$. This new CBS will select a search node from $Focal$ to expand, instead of that from $Open$. Compared to the original CBS, the new algorithm also performs the best-first search on $Focal$, but the search priority changes from the solution cost to a new heuristic function $\psi$. Typically, $\psi$ is a handcrafted function that takes a solution as the input, and outputs a value that prefers solutions with fewer conflicts. We instead use a learned heuristic function based on the {\bf Graph Transformer}.

\begin{algorithm}[!h]
\begin{algorithmic}
\NoThen
\NoDo
\caption{CBS with Biased Heuristics~\cite{barer2014suboptimal, huang2021learningb}}
\label{alg:focal_search}
\STATE {\bf Input:} A MAPF instance and suboptimality factor $w$
\STATE {\bf Input:} Heuristic function $\psi$ (e.g., Graph Transformer)
\STATE Generate the root search node $R$ with an initial solution
\STATE Initialize open list $Open \leftarrow \{R\}$
\STATE $LB \leftarrow c(R)$, and initialize focal list $Focal \leftarrow \{R\}$
\WHILE{$Open$ is not empty}
\STATE $N^* \leftarrow$ $\arg\min_{N \in Focal} \psi(\sigma(N))$
\STATE $C \leftarrow$ first conflict in $\sigma(N^*)$
\IF{$C$ does not exist}
{\STATE {\bf return} solution $\sigma(N^*)$ }
\ENDIF
\STATE Remove $N^*$ from $Open$ and $Focal$
\IF{$\min_{N\in Open} c(N) > LB$}
\STATE $LB = \min_{N\in Open} c(N)$
\STATE $Focal = \{N \in Open : c(N) \leq w\cdot LB \}$
\ENDIF
\STATE Generate two children nodes $N_1$ and $N_2$ from node $N^*$\\
\STATE Add $C_i$ to $\mathcal{T}(N_i)$, for $i = 1, 2$\\
\STATE Call low-level planner to get $\sigma(N_i)$, for $i = 1, 2$ 
\STATE Add $N_i$ to $Open$, for $i = 1, 2$
\STATE Add $N_i$ to $Focal$ if $c(N_i) \leq w\cdot LB$, for $i = 1, 2$
\ENDWHILE
\STATE {\bf return} No solution
\end{algorithmic}
\end{algorithm}

{\bf Proposition.} {\em Algorithm \ref{alg:focal_search} is complete and bounded-suboptimal with a factor of $w\geq 1$, as mentioned in~\cite{barer2014suboptimal}.}

Suppose the problem is feasible, but Algorithm \ref{alg:focal_search} does not find a solution given a sufficient time budget. Then for an arbitrary search node $N$ with a feasible solution, there exists an ancestor search node $N^p$ added to $Open$ but not expanded. Suppose $N^{p*}$ is the search node with the lowest cost among these unexpanded ancestors. $N^{p*}$ is not selected by $Focal$, since it is not expanded. Thus, either (i) $N^{p*}$ is not in $Focal$, or (ii) $N^{p*}$ is in $Focal$ but not selected. (i) is impossible, because there do not exist infinitely many solutions that have costs lower than $\frac{1}{w}\cdot c(N^{p*})$. (ii) cannot happen, because there do not exist infinitely many solutions with costs lower than or equal to $w\cdot c(N^{p*})$. As a result, $N^{p*}$ will be selected eventually and expanded. Therefore, we have proved the algorithm to be complete by contradiction. The focal search never expands search nodes with costs higher than $w$ times the optimal solution; therefore, it is bounded-suboptimal with a factor of $w$.

We note that the focal search described here is a special case of BCBS~\cite{barer2014suboptimal}, i.e., BCBS ($w, 1$), as the focal search is only applied to the high-level planner. In our graph-based problem settings, there is no significant improvement when applying the focal search to the low-level planner. Rather, if we introduce the focal search to the low-level planner, it would consume a notable portion of computation on the collision checking of edges, which has no improvement in the overall performance. We refer readers to Question 4 in Section \ref{sec:ablation_study}  for further details.

\section{Graph Transformers as Heuristic Functions}

In this section, we describe the architecture of the Graph Transformer  and how to train it represent a heuristic function that accelerates CBS.

\begin{figure}[h]
  \centering
  \includegraphics[width=0.41\textwidth]{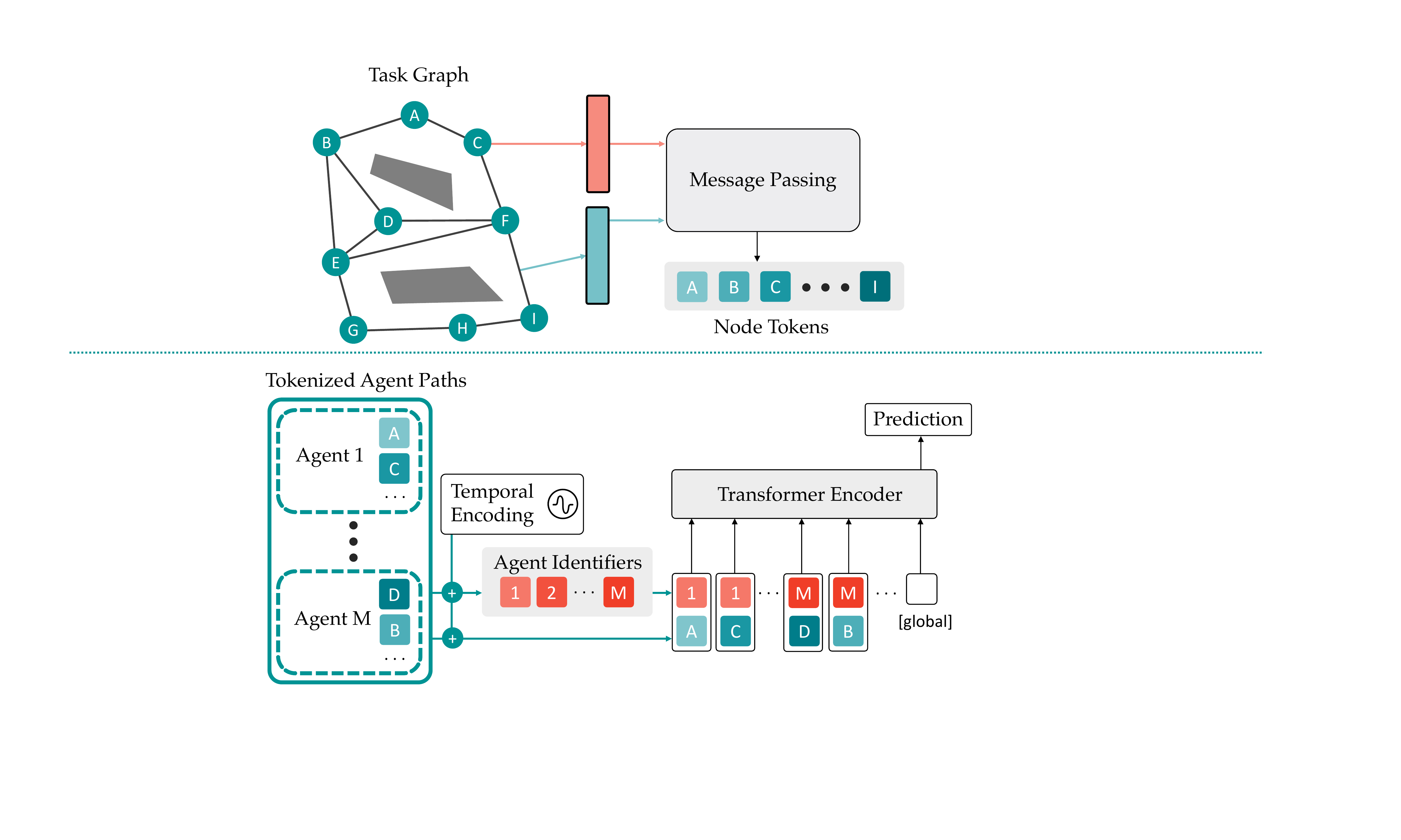}
  \caption{The proposed Graph Transformer architecture. It has several desired properties that are specifically designed to deal with MAPF inputs. See Section \ref{sec:property_arch} for more details.}
  \label{figurelabel}
  \vspace{-15pt}
\end{figure}

\subsection{Network Architecture}

The input to the graph transformer $\phi$ is a graph $G$, and a solution $\sigma=\{\sigma_{i}\}_{i\in[1\cdots M]}$. The output $\phi(G, \sigma)$ predicts a scalar value as the heuristic. Such a predicted value correlates with the chance that the current search node will yield descendant search nodes with feasible solutions. For example, compared to those nodes that cannot eventually reduce the conflicts, the promising nodes leading to feasible solutions should have lower $\phi(G, \sigma)$ values. 

The graph transformer has two stages: {\em graph tokenization} and {\em attentive aggregation}. We describe each stage as follows.

{\bf Stage 1: Graph Tokenization.} The graph tokenization transforms each graph vertex into an embedding using a Graph Neural Network (GNN). Here we use Message-Passing Neural Networks~\cite{mpnn} (MPNN) as the GNN architecture. The input to the MPNN is a graph $G=\langle V, E\rangle$, where the feature $d^v_i$ for each graph vertex $v_i\in V$ is its respective 2D position, and the feature $d^e_l$ for each edge $e_l=(v_i\rightarrow v_j)$ is the relative position of $v_j$ to $v_i$. With two linear layers $f_x$ and $f_y$, the vertices and edges are first encoded as $x$ and $y$ using $\forall v_i\in V, x_i=f_x(d_i^v); \forall e_l \in E, y_l=f_y(d_l^e)$. Then, using three MLPs $\{f_k\}_{k\in [1,2,3]}$, the MPNN updates the information for each graph vertex $v_i \in V$ as follows:
\begin{align}
x_i\leftarrow x_i+\max\{f_k(x_i, x_j, y_l)\},  \forall e_l:(v_i\rightarrow v_j)\in E.
\end{align}

After all $x_i$ are updated using MLP $f_1$, the MPNN continues to update $x_i$ using MLP $f_2$ and so on. The $\max$ denotes the max-pooling over the feature dimension. We use max-pooling to take a set with an arbitrary number of elements while ensuring robustness~\cite{pointnet}. Since it is invariant to the permutation of these elements, the MPNN here can take graphs with an arbitrary number of vertices and edges, but also is permutation invariant by construction.

{\bf Stage 2: Attentive Aggregation.} After we compute the token $x_i$ for each graph vertex $v_i$ from Stage 1, we model the inter-agent interactions using the Transformer~\cite{vaswani2017attention}. The path of each agent $\sigma_i$ is first tokenized as $\rho_i=\{x_j, \forall v_j \in \sigma_i\}$. To inject the temporal information of these tokenized solutions, we introduce Temporal Encoding~\cite{DBLP:journals/corr/abs-2210-08408}. The approach is similar to \cite{vaswani2017attention, mildenhall2021nerf} (as positional encoding in their settings). We denote $\rho_i^t\in\mathbb{R}^{D}$ as the vertex token of agent $i$ at time step $t$. For each token $\rho_i^t$, we add it element-wisely with a temporal encoding $\rho_i^t\leftarrow \rho_i^t+TE(t) \in \mathbb{R}^{D}$. The $2k$-th and $2k$+$1$-th dimensions of $TE(t)$ are as follows:
\begin{align}
    TE(t)_{2k} = \sin(t/{10000^{2k/{D}}}),\\ 
    TE(t)_{2k+1} = \cos(t/{10000^{2k/{D}}}).
    \label{eq:TE}
\end{align}

The hyperparameter 10000 is used following the common practice~\cite{vaswani2017attention}. To encourage the model to be aware of which agent each token $\rho_i^t$ belongs to, we concatenate the Agent Identifier $\tau_i$ to each token $\rho_i^t\leftarrow \rho_i^t || \tau_i$, similar to \cite{kim2022pure}. For each agent $i$, the agent identifier $\tau_i$ is calculated by taking the max-pooling over all its vertex tokens: $\tau_i = \max\{\rho_i^t\}, \forall t \in [1\cdots T_i]$. Then, all tokens from the solution of all agents $\{\rho_i^t: \forall i\in [1\cdots M],\forall t\in[1\cdots T_i]\}$ will be fed as the input to the Transformer Encoder. For global prediction, we append an extra trainable token $global$ to the input, following the common practice \cite{devlin2018bert,dosovitskiy2020image}. The Transformer Encoder predicts an output for each input token, and we use the output of the trainable $global$ token as $\zeta$. With a linear layer $f_\phi: \mathbb{R}^D\rightarrow \mathbb{R}$, the final output is computed as $\phi(G, \sigma)=f_\phi(\zeta)$.

Here we use the Transformer Encoder, since it enables the Graph Transformer to take a variable number of tokens and model their dependencies, while preserving the invariance to the permutation of tokens. We refer readers to \cite{vaswani2017attention} for more details on the Transformer Encoder.

{\bf Properties of the Graph Transformer.} \label{sec:property_arch} By construction, the Graph Transformer is able to handle the input graph with a variable number of vertices and edges, agents with a variable total number, and the input solution with a variable length. Additionally, it is aware of the temporal information, and inter-agent interactions. Its output is permutation invariant to both the orders of graph vertices and the agents.
\subsection{Training Graph Transformers} 
\begin{figure}[h]
  \centering
  \includegraphics[width=0.48\textwidth]{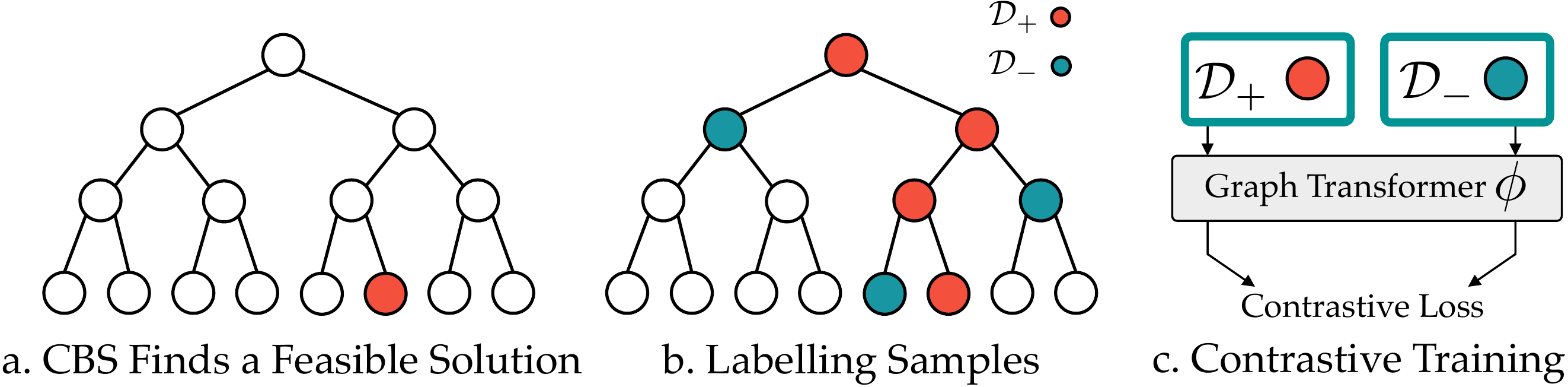}
  \caption{The training framework. We use a supervised Contrastive Loss. The labels are generated from the CBS search tree.}
  \label{figurelabel}
  \vspace{-10pt}
\end{figure}

\begin{figure*}[!h]
  \centering
  \includegraphics[width=1.0\textwidth]{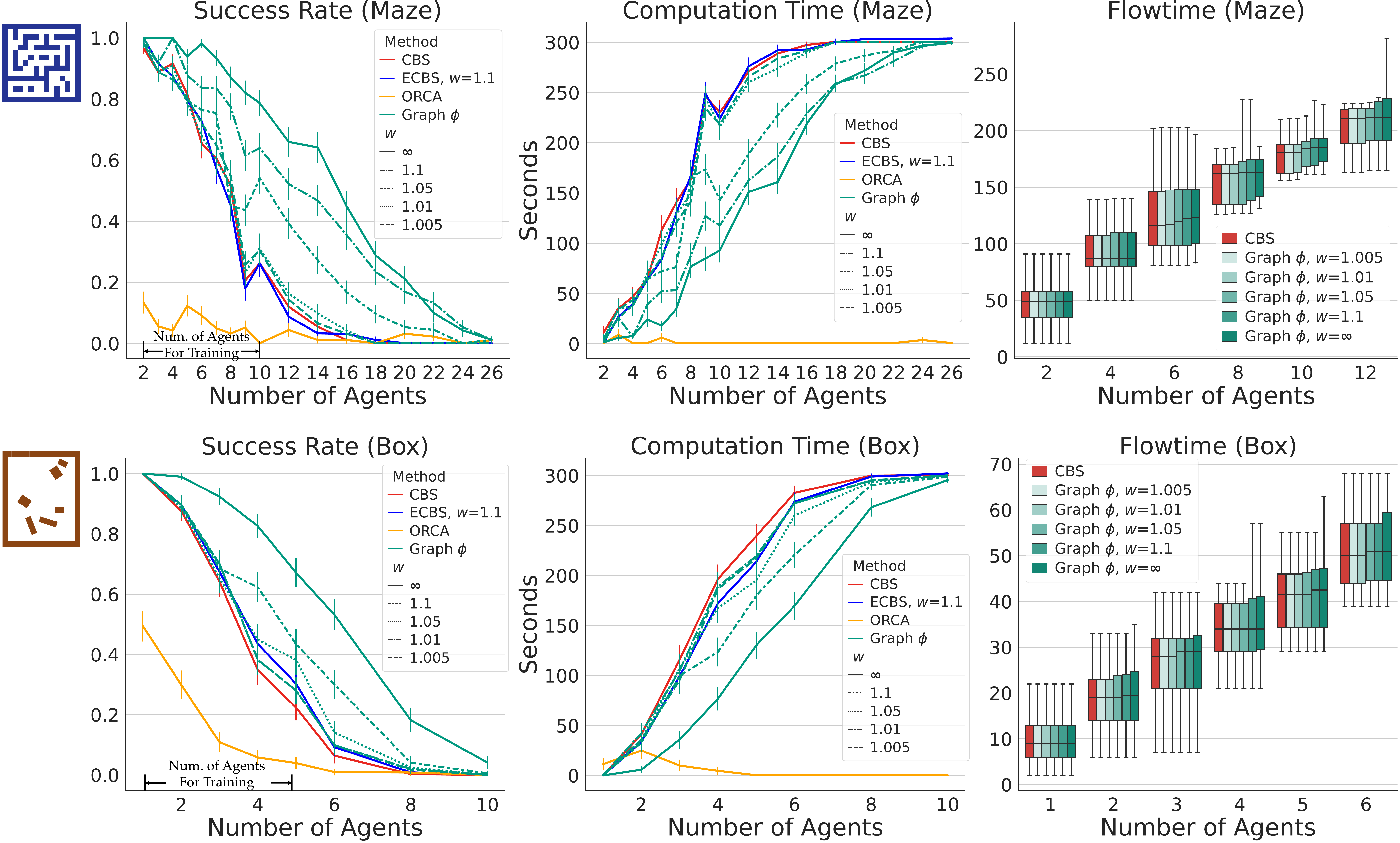}
  \caption{Success rates, computation time, and flowtime within the runtime limit of 5 minutes, as functions of the number of agents. The results are averaged over 100 test instances for each setting of the agent number. We evaluate our approach with $w\in [1.005,1.01,1.05,1.1,\infty]$, and compare its performance with CBS, ECBS ($w=1.1$) and ORCA. Though trained with relatively few agents, results have shown that our approach generalizes well and significantly outperforms the baselines.}
  \label{fig:overall_performance}
  \vspace{-15pt}
\end{figure*}

\noindent{\bf Data Generation.} Given a MAPF instance, we first use CBS to generate feasible solutions. No data will be collected if CBS fails to solve the instance. If it succeeds, we start to collect positive and negative samples from its search tree. A positive sample will be collected by dataset $\mathcal{D}_+$, if itself or one of its descendant search nodes contains the feasible solution. A negative sample will be collected by dataset $\mathcal{D}_-$, if it is a sibling search node of a positive sample. We record each sample's graph $G$ and solution $\sigma$. In addition, we record the value $d$ as its respective depth in the search tree.

\noindent{\bf Supervised Contrastive Learning.} The objective of the model is to learn a ranking of the samples. Namely, given an arbitrary pair of positive sample $(G, \sigma_+, d_+)$ and a negative sample $(G, \sigma_-, d_-)$ from the same MAPF graph $G$, we learn the following ranking:
\begin{align*}
\phi(G, \sigma_+) < \phi(G, \sigma_-), \text{ if } d_+ \geq d_-.
\end{align*}

We illustrate the intuition here. Imagine such ranking is learned perfectly and $w=\infty$, meaning all the leaf search nodes will be in $Focal$. Suppose at some time point, $Focal$ includes 1 positive sample $p_+$. $Focal$ may or may not include negative samples. If they exist, then their depths are no deeper than $d(p_+)$. Define this condition as a loop invariant. Then $p_+$ will be selected and expanded first according to the ranking. If $p_+$ is the feasible solution, then the algorithm terminates. Otherwise, $Focal$ will have one positive sample $p'_+$ and some negative samples, and all negative samples have depths no deeper than $d(p'_+)$. Thus, the loop invariant remains true. The loop invariant is also true for the base case, where $Focal$ only has the root search node. Therefore, by learning such ranking, we encourage Algorithm \ref{alg:focal_search} to expand positive samples first and expand the negative samples as few as possible, which could save significant computation and greatly accelerate CBS.

We use supervised contrastive learning to train a Graph Transformer $\phi$ that ranks the positive samples above the negative samples. Given an arbitrary pair of positive and negative samples, $p_+: (G_+, \sigma_+, d_+)\in \mathcal{D}_+, p_-: (G_-, \sigma_-, d_-)\in \mathcal{D}_-$, this pair is defined to be valid as: $\mathbb{I}(p_+, p_-): (G_+=G_-) \wedge (d_+ \geq d_-)$. With a hyperparameter $\gamma=0.1$, we define $\delta(x): \max(0, \gamma + x)$. We aim to minimize the Contrastive Loss as follows:
\begin{align}
\frac{1}{L}\sum_{\substack{p_+\in \mathcal{D}_+ \\ p_-\in \mathcal{D}_-}}\delta(\phi(G_+, \sigma_{+})-\phi(G_-, \sigma_-))\cdot \mathbb{I}(p_+, p_-),
\end{align}
where $L=\sum^{p_+\in \mathcal{D}_+}_{p_-\in \mathcal{D}_-}\mathbb{I}(p_+, p_-)$ is the total number of valid sample pairs, and $\mathcal{D}^+, \mathcal{D}^-$ are the datasets of positive and negatives samples respectively.

Once the training of Graph Transformer $\phi$ reaches convergence, we could deploy it as the heuristic function $\psi$ in Algorithm \ref{alg:focal_search}. However, in practice, we found that the model would predict relatively low values for multiple search nodes in the focal set, indicating that all of them may lead to feasible solutions. To break the tie, we instead represent $\psi$ as the depth $d$ combined with $\phi$, i.e., $\psi=\langle -d, \phi\rangle$. It means that the algorithm would first prefer the search nodes with deeper depths; if there exist multiple search nodes with the deepest depths, then it would prefer the search nodes with lower $\phi$. Such a design enables the algorithm to make decisions consistently if multiple promising nodes exist. Without further specification, we denote our approach with such a heuristic as {\bf Graph $\phi$}.

\section{Experiments} 

\subsection{Main Experiments} 


\begin{figure*}[!th]
  \centering
  \includegraphics[width=1.0\textwidth]{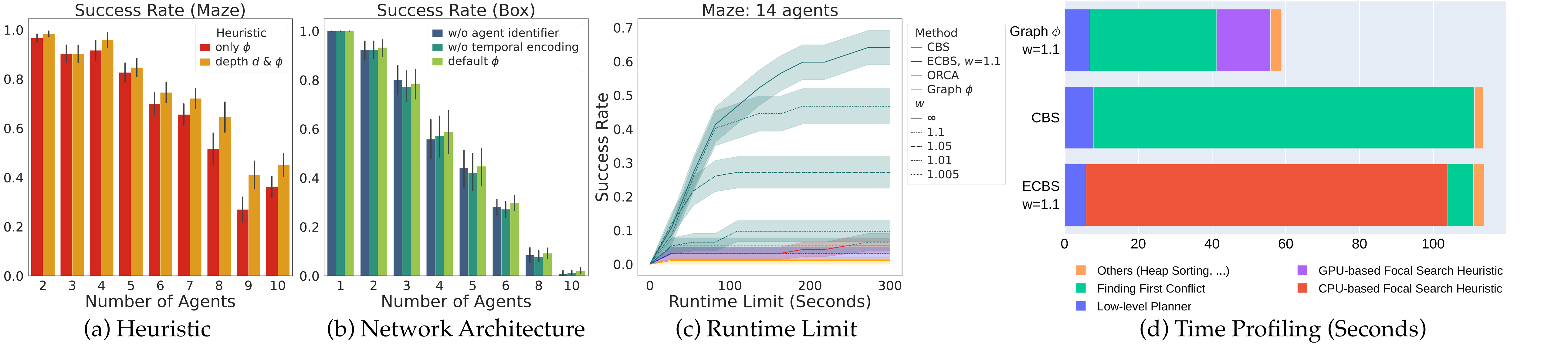}
  \caption{We conduct 4 various ablation studies to evaluate the proposed method systematically. See Section~\ref{sec:ablation_study} for more details.}
  \label{fig: ablation_study}
  \vspace{-20pt}
\end{figure*}

\noindent{\bf Experimental Setup.}
We design two types of environments for evaluation: Maze and Box (see Fig.~\ref{fig: problem_instance}). Each environment includes 2700 MAPF instances with samples generated by CBS for training. For testing, there are 100 MAPF test instances w.r.t. each agent number setting. We generate a random map for each Maze instance and a set of random obstacles for each Box instance. The graph and start and goal vertices are also generated randomly for each instance.

The training instances vary between 2 and 10 agents for Maze, and vary between 1 and 5 agents for Box. The test instances vary between 2 and 26 agents (2-10, 12, 14, 16, 18, 20, 22, 24, 26) for Maze, and vary between 1 and 10 agents (1-5, 6, 8, 10) for Box. We ensure that the test instances are unseen in the training set. All experiments were conducted using a 12-core, 3.2Ghz i7-8700
CPU and 4 Nvidia GTX 1080Ti GPUs. We test $w\in [1.005,1.01,1.05,1.1,\infty]$ for our method. For all methods, we set the runtime limit as 5 minutes, following the common practice~\cite{huang2021learningb}.

\noindent{\bf Baselines.} We compare our method with 3 baselines: (i) Conflict-Based Search. (ii) Enhanced Conflict-Based Search (ECBS): a bounded-suboptimal version of CBS~\cite{barer2014suboptimal}. It applies focal search to both high-level and low-level planners, using hand-crafted heuristic functions. We choose its $w=1.1$. (iii) ORCA
~\cite{vandenberg_Reciprocal_2008}: a reactive collision avoidance algorithm, which works effectively in low density environments. 

We find that there are very few open-source implementations that could directly apply CBS and ECBS to non-grid-based problems. As a result, we implement all tree-search methods (CBS, ECBS, and our method) from scratch using Python. We use basically the same framework when implementing CBS, ECBS, and our approach. To make the comparison fair, we ensure that all these 3 methods are aggressively optimized by strictly following the C++ implementations \footnote{https://github.com/whoenig/libMultiRobotPlanning/} and original paper~\cite{sharon_Conflictbased_2015, barer2014suboptimal}, and using Bayesian hyperparameter search for the $w$ of ECBS.

\noindent{\bf Evaluation Metrics.} Our evaluation includes 3 metrics: 1) {\em Success Rate}, the ratio of the number of successful instances to the total number of test instances. An instance is successful if all agents reach their goals with no collision before the timeout happens.
2) {\em Computation Time}, the average computational time for each instance, including the failed ones. 
\footnote{Since CBS and ECBS call different low-level planners (A* and A*-$\epsilon$), we consider the computation time to be the fairest metric to evaluate the efficiency, instead of counting the number of expanded nodes, for instance.} 3) {\em Flowtime}, the sum of all the agents' travel time. We only compare the flowtime of the CBS to our method based on the instances where both methods succeed, to show our approach's bounded-suboptimality guarantees.

\noindent{\bf Overall Performance.} We demonstrate the overall performance in Figure \ref{fig:overall_performance}. Our method significantly outperforms the three baselines in both Maze and Box. It requires much less computation time and achieves significantly higher success rates. Furthermore, though only trained with relatively few numbers of agents, our method generalizes remarkably well to a higher number of agents. For example, our network trained by 2 to 10 agents can be generalized up to 26 agents in Maze, while our policy trained by 1 to 5 agents can be generalized up to 10 agents in Box. In particular, with $w=1.1$, our method achieves the success rates of $47\%, 23\%, 10\%, 1\%$ in Maze with 14, 18, 22, and 26 agents, and achieves the success rates of $62\%, 30\%. 4\%, 0.5\%$ in Box with 4, 6, 8, and 10 agents. On the other side, CBS fails to solve Maze with 18 agents and Box with 10 agents within the timeout. Similarly, ECBS starts to fail in all instances with 20 agents for Maze and with 10 agents for Box. In addition, since the bounded-suboptimality of our algorithm is proved theoretically, it is not surprising to see that the solution qualities (flowtime) of our method are very close to the optimal solutions. Finally, ORCA easily fails in highly dense environments, and such performance is consistent with previous works~\cite{DBLP:journals/ral/ArulSPOXLM19}.

\subsection{Ablation Study}\label{sec:ablation_study}

In this section, we investigate four questions:

{\bf (a) Is incorporating depth information into the heuristic beneficial to the performance?}
Fig.~\ref{fig: ablation_study} (a) illustrates the comparison between the heuristic with only $\phi$, and with depth $d$ and $\phi$, on tests from 2 to 10 agents in Maze. The result shows the effectiveness of introducing depth information. The improvement becomes more noteworthy as the number of agents grows, which validates our choice.

{\bf (b) Do the Temporal Encoding and the Agent Identifier improve the performance?}
Fig.~\ref{fig: ablation_study} (b) demonstrates the performances in Box with and without the Temporal Encoding and Agent Identifier. The results show that Agent Identifier and Temporal Encoding improve the success rate. The average improvement of introducing Agent Identifier and Temporal Encoding over the non-default settings is 16$\pm$33\%. 

{\bf (c) How will the runtime limit affect the performance?}
We take the tests with 14 agents in Maze as an example. We set different runtime limits from 25 seconds to 300 seconds with the interval as 25 seconds. In Fig.~\ref{fig: ablation_study} (c), we show that our methods outperform the baselines regardless of how the runtime limit changes. When the limit is 300 seconds, our method achieves a success rate of 64\% ($w=\infty$), while CBS, ECBS, and ORCA only have 5\%, 3\%, and 1\% respectively.

{\bf (d) Time profiling each module of the planners.} Finally, we wish to answer the question of why the performances of ECBS considerably degrade once we apply it to non-grid-based MAPF instances, compared to the traditional grid-based settings. We profile each method, and average the results over the Maze tests with 2-10 agents. 

Fig.~\ref{fig: ablation_study} (d) illustrates that ECBS spends considerable computation on the focal heuristic calculation. Such behavior is reasonable, since now for dense graphs, the heuristic is calculated by checking the collisions along edges. Such collision checking is often the main bottleneck for planning and by itself NP-hard in general~\cite{chazelle1984convex,jimenez1998collision}. Meanwhile, our method uses a learned function for the focal heuristic calculation (the GPU part), which only takes 15\% computation cost compared with ECBS. Compared with CBS and ECBS, our method only requires 50\% of the total computation time. 

\vspace{-0.4em}

\section{Conclusion and Future Work}
 
We proposed the Graph Transformer as a heuristic function to accelerate CBS by guiding the tree search towards promising nodes with feasible solutions. While achieving significant acceleration, our approach guarantees provable completeness and bounded-suboptimality. We show that in two continuous environments with dense motion graphs, our method outperforms three classical MAPF baselines (CBS, ECBS, and ORCA), while generalizing to unseen tests with a higher number of agents remarkably well. Future works include real-world experiments and improving the method for 10x-100x agents.

~\\
\noindent{\bf Acknowledgment.} This material is based on work supported by ARL DCIST CRA W911NF-17-2-0181, European Research Council (ERC) Project 949940 (gAIa), the United States Air Force and DARPA under Contract No. FA8750-18-C-0092, AFOSR YIP FA9550-19-1-0041, NSF Career CCF 2047034, Amazon Research Award, and a gift through Amazon.com Inc.

\IEEEpeerreviewmaketitle

\bibliographystyle{IEEEtran}
\bibliography{reference}

\end{document}